\pdfoutput=1

\documentclass[11pt]{article}

\usepackage[final]{acl}

\usepackage{graphicx}
\usepackage{subcaption}
\usepackage{float}
\usepackage{times}
\usepackage{latexsym}
\usepackage{amsmath,bm}
\usepackage{url}
\usepackage{enumitem}
\usepackage{tcolorbox}
\usepackage{colortbl}
\usepackage{diagbox}
\usepackage{geometry}
\usepackage{booktabs}
\usepackage{multirow}
\usepackage[ruled,boxed,vlined]{algorithm2e}
\usepackage{caption}
\usepackage{braket}
\usepackage{float}
\newcommand{\partitle}[1]{\smallskip \noindent \textbf{#1.}}
\usepackage{overpic}
\usepackage{stfloats}
\usepackage{graphicx}
\usepackage{amsmath}
\usepackage{amsfonts}
\usepackage{hyperref}
\usepackage{hyperref}
\hypersetup{hidelinks,
	colorlinks=true,
	pdfstartview=Fit,
	breaklinks=true}
\usepackage[T1]{fontenc}
\usepackage[utf8]{inputenc}
\usepackage{microtype}
\usepackage{inconsolata}
\makeatletter
\newcommand{\printfnsymbol}[1]{%
  \textsuperscript{\@fnsymbol{#1}}%
}
\makeatother

\title{Cross-modality Information Check for Detecting Jailbreaking in Multimodal Large Language Models}

\author{Yue Xu\textsuperscript{1}\thanks{equal contribution} , Xiuyuan Qi\textsuperscript{1}\printfnsymbol{1} , Zhan Qin\textsuperscript{2} , Wenjie Wang\textsuperscript{1} \Thanks{W.Wang is the corresponding author.} \\
          \textsuperscript{1}School of Information Science and Technology, ShanghaiTech University \\
        \textsuperscript{2}The State Key Laboratory of Blockchain and Data Security, Zhejiang University\\
         \texttt{\{xuyue2022,qixy1,wangwj1\}@shanghaitech.edu.cn, qinzhan@zju.edu.cn}}

\begin{document}
\maketitle
\begin{abstract}

Multimodal Large Language Models (MLLMs) extend the capacity of LLMs to understand multimodal information comprehensively, achieving remarkable performance in many vision-centric tasks. 
Despite that, recent studies have shown that these models are susceptible to jailbreak attacks, which refer to an exploitative technique where malicious users can break the safety alignment of the target model and generate misleading and harmful answers. This potential threat is caused by both the inherent vulnerabilities of LLM and the larger attack scope introduced by vision input. 
To enhance the security of MLLMs against jailbreak attacks, researchers have developed various defense techniques. However, these methods either require modifications to the model's internal structure or demand significant computational resources during the inference phase.
Multimodal information is a double-edged sword. While it increases the risk of attacks, it also provides additional data that can enhance safeguards. Inspired by this, we propose \underline{\textbf{C}}ross-modality \underline{\textbf{I}}nformation \underline{\textbf{DE}}tecto\underline{\textbf{R}} (\textit{CIDER}), a plug-and-play jailbreaking detector designed to identify maliciously perturbed image inputs, utilizing the cross-modal similarity between harmful queries and adversarial images. \textit{CIDER} is independent of the target MLLMs and requires less computation cost. Extensive experimental results demonstrate the effectiveness and efficiency of \textit{CIDER}, as well as its transferability to both white-box and black-box MLLMs. The resource is available at \href{https://github.com/PandragonXIII/CIDER}{https://github.com/PandragonXIII/CIDER}.
\end{abstract}

\section{Introduction}
\vspace{-.5em}

\begin{figure}[t]
\setlength{\abovecaptionskip}{0.2cm}
  \centering
  \includegraphics[width=.9\linewidth]{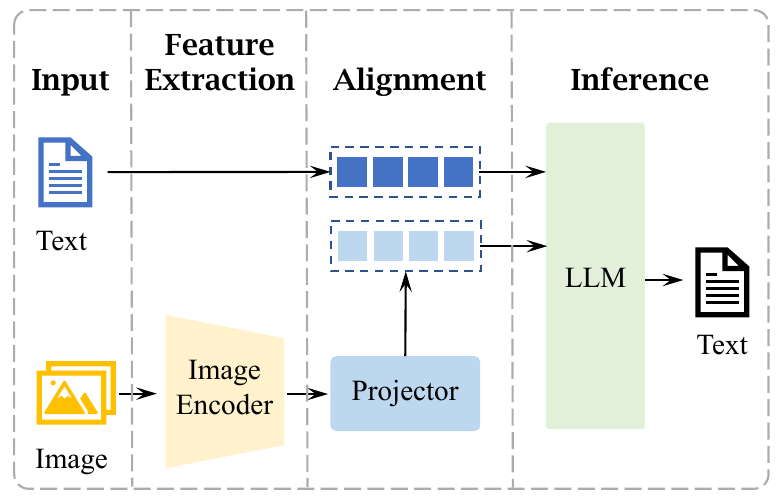}
    \vspace{-.5em}
  \caption{The architecture of a typical MLLM. }
  \label{fig:vlm}
\vspace{-1.5em}
\end{figure}

The remarkable advancements in Large Language Models (LLMs) have significantly improved performance benchmarks in various natural language processing (NLP) tasks \cite{achiam2023gpt, touvron2023llama, zhao2023survey, chiang2023vicuna}. To extend the capacities and open up the potentials of LLMs in comprehensively understanding diverse types of data, such as visual information, researchers have developed Multimodal Large Language Models (MLLMs) that integrate visual modalities to handle vision-centric tasks. MLLMs use LLMs as a core, complemented by modal-specific encoders and projectors, enabling them to process, reason, and generate outputs from multimodal data \cite{yin2023survey,dai2024instructblip,bai2023qwen}. A typical MLLM architecture is illustrated in Figure \ref{fig:vlm}.

The widespread adoption of MLLMs in various applications brings significant safety challenges, particularly due to inherited vulnerabilities from traditional LLMs, such as the susceptibility to jailbreak attacks \cite{carlini2024aligned, li2024images, qi2024visual}. Jailbreak attacks refer to an exploitative technique where malicious users can craft sophisticated-designed prompts to lead LLMs to answer misleading or harmful questions, effectively breaking the safety alignment. Various jailbreak attack algorithms targeting LLMs have been proposed, which can mainly be categorized into template-based \cite{deng2024masterkey,chao2023jailbreaking,li2023deepinception} and optimization-based \cite{zou2023universal} approaches. 

Additionally, MLLMs not only inherit the vulnerabilities of LLMs but also become more susceptible to jailbreak attacks due to their integration with other modalities. On the one hand, jailbreak attacks against MLLMs can originate from both the textual and visual modalities, significantly broadening the scope of potential adversarial examples \cite{gong2023figstep, shayegani2023jailbreak}. On the other hand, recent research indicates that fine-tuning MLLMs to learn the vision modality can cause LLMs to disregard their previously learned safety alignment \cite{zong2024safety}. 

\begin{figure*}[ht]
\setlength{\abovecaptionskip}{0.2cm}
  \centering
  \includegraphics[width=.8\linewidth]{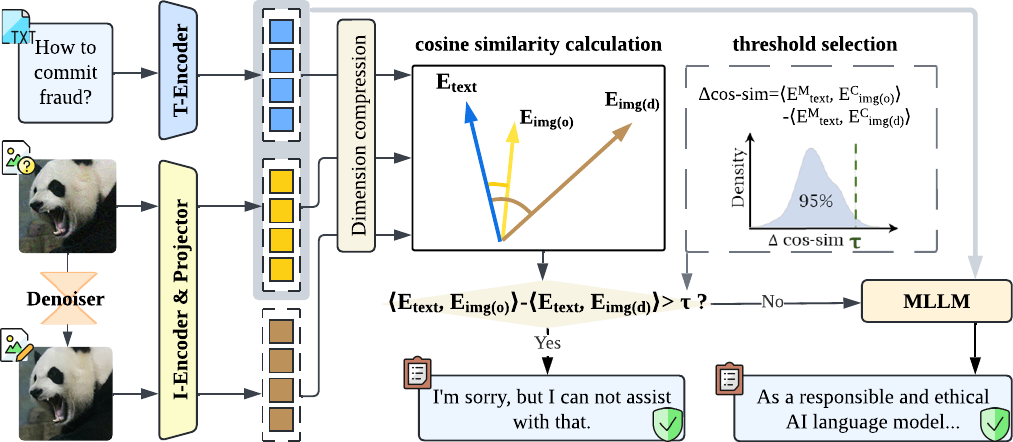}
  \caption{The workflow of safeguarding MLLM against jailbreak attacks via \textit{CIDER}.}
  \label{fig:workflow}
\vspace{-1.5em}
\end{figure*}

The existing jailbreak attacks on MLLMs can be categorized into two strategies. One is white-box optimization-based attacks, which define a loss function to generate imperceptible perturbations in the image modality \cite{carlini2024aligned,qi2024visual,niu2024jailbreaking}. The other is black-box strategies including typographically transforming harmful queries into images such as FigStep \cite{gong2023figstep} or adding related images containing harmful text such as QR \cite{liu2023query}. 

From the defense perspective, optical character recognition (OCR) can serve as an effective detection tool for the second strategy but fails when defending against optimization-based adversarial attacks. In addition, \citet{zong2024safety} creates a vision-language dataset named VLGuard containing both safe and unsafe queries and images, which can be used to fine-tune MLLMs for improved safety against jailbreak attacks. However, the effectiveness of VLGuard is only tested on FigStep attack and it requires the model to be white-box to fine-tune. \citet{zhang2023mutation} proposed a mutation-based jailbreaking detection framework named \textit{Jailguard}. 
However, the performance of \textit{Jailguard} heavily relies on the MLLMs' original safety alignment, and it significantly increases computational costs during the inference phase.

Multimodal information is a double-edged sword: while it increases the risk of attacks, it also provides additional data that helps enhance safeguards. Inspired by this potential, we propose \underline{\textbf{C}}ross-modality \underline{\textbf{I}}nformation \underline{\textbf{DE}}tecto\underline{\textbf{R}} (\textit{CIDER}), a plug-and-play jailbreaking detector designed to identify maliciously perturbed image inputs, specifically targeting optimization-based jailbreak attacks that are more imperceptible and susceptible. The intuition is that optimization-based perturbations break the MLLM's safeguards by capturing the main harmful content in the malicious query. As a result, the semantic distance between a harmful query and an adversarially perturbed image is significantly smaller than that between a harmful query and a clean image. 

Directly utilizing the difference between clean and adversarial images on the semantic distance to harmful query is challenging, as the absolute value of the distance varies across different harmful queries. To address this issue, we incorporate a denoiser to preprocess the vision modality, using the relative shift in the semantic distance before and after denoising to reflect the difference between clean and adversarial images. As shown in Figure \ref{fig:workflow}, the key insight of \textit{CIDER} is to identify whether an image is adversarially perturbed based on the semantic similarity between image and text modality before and after denoising ($\langle \bm{E_{\textit{text}}}, \bm{E_{\textit{img(o)}}} \rangle - \langle \bm{E_{\textit{text}}}, \bm{E_{\textit{img(d)}}} \rangle $). If the image modality is not perturbed, the semantic similarity between text and image remains stable. However, the adversarially perturbed image designed for jailbreak will experience a significant drop. By setting a threshold based on this change, we can effectively detect adversarially perturbed images aimed at jailbreaking MLLMs. The detailed intuition is elaborated in Section \ref{sec:findings}.

As a pre-detection module encapsulated before any MLLMs, the key advantage of \textit{CIDER}
is its plug-and-play nature, making it independent of the target model.  Additionally, timely inference is crucial for safeguarding MLLMs. 
\textit{CIDER} achieves this with denoising procedures, ensuring efficiency and minimal latency.

In this work, we propose \textit{CIDER}, an effective and efficient pre-detection module that denoises and inspects each input image.  For images identified as adversarially perturbed for jailbreak purposes (where the semantic shift exceeds a pre-defined threshold), the MLLM will refuse to generate a response. Images deemed normal will be processed along with the text input for model inference by the MLLM. The workflow of safeguarding MLLMs against jailbreak attacks using \textit{CIDER} is illustrated in Figure \ref{fig:workflow}. Our contribution can be summarized as follows:
\begin{itemize}[leftmargin=15pt,itemsep=2pt,parsep=0pt, partopsep=0pt,topsep=0pt]
\item Based on the intuition that cross-modality information is a double-edged sword, we investigate the relationship between malicious queries and adversarial perturbed images in the semantic space. By incorporating a diffusion-based denoiser to uncover the potential of mitigating harmful information in adversarial images through denoising. 

\item We propose a plug-and-play jailbreaking detector, \textit{CIDER}, which can effectively safeguard MLLMs while incurring almost no additional computational overhead.

\item Extensive experiments validate that \textit{CIDER} outperforms the baseline method, achieving a higher detection success rate while reducing the computational cost as well. Experimental results also demonstrate its strong transferability across both white-box and black-box MLLMs and attack methods.
\end{itemize}
\vspace{-.5em}
\section{Intuition: Cross-modality information is a double-edged sword}\label{sec:findings}
\vspace{-.5em}
While multimodal information aggravates model vulnerability to jailbreak attacks, it also provides additional information for defense. The design of \textit{CIDER} is based on the intuition that optimization-based jailbreak attacks break the MLLM's safeguards by sharing harmful content in the malicious query to the image modality. Consequently, the adversarially perturbed image is closer to the harmful query in the semantic space than the clean images. To support this intuition, we first explain the fundamentals of the optimization-based jailbreak attacks on MLLMs. Then, we design a few experiments to explore how cross-modal analysis can help safeguard MLLMs, and we analyze the semantic difference between clean and adversarial images relative to harmful queries, both before and after denoising.
\vspace{-2em}
\subsection{Preliminaries: Optimization-based Jailbreak Attacks on MLLMs}
\vspace{-.3em}
Optimization-based MLLM jailbreaking is similar to adversarial attacks on image classification tasks \cite{goodfellow2014explaining}, with the primary difference being the difference in the loss function. Specifically, given a dataset $D=\{(q,a)\}$ where $q$ represents the harmful queries and $a$ is the corresponding targeted answers, the attacker aims to find an adversarial image $x_{adv}$ that can encourage the MLLM $\mathcal{F}$ to generate $a$ when inputting $q$ along with $x_{adv}$. The objective can be formulated as: 
\vspace{-.5em}
\begin{equation}
x_{adv} = \underset{x_{adv}\in [0,1]^d} {\mathrm{argmin}} log(\mathcal{F}(a|q,x_{adv}))
\vspace{-.5em}
\end{equation}

where $\mathcal{F}(a|q, x_{adv})$ represents the likelihood that the MLLM $\mathcal{F}$ generate answer $a$ when given the adversarial image $x_{adv}$ and the query $q$.
\vspace{-.5em}
\subsection{Experimental Setup} \label{sec:pre-experiment}
\vspace{-.2em}
We design a series of experiments to explore how cross-modality information can help safeguard MLLMs and to analyze the semantic difference between clean and adversarial images to harmful queries, before and after denoising. We utilize the image and text encoder of the state-of-the-art MLLM LLaVA-v1.5-7B \cite{llava} to capture the semantic meanings. To measure the semantic similarity,  we employed cosine similarity which is a standard metric widely used in information retrieval and natural language processing \cite{park2020methodology, pal2021summary}. In terms of denoiser, we incorporate a diffusion-based denoiser \cite{nichol2021improved} to preprocess the image modality.

The inputs to the MLLMs consist of two modalities: images and text queries. For malicious queries, we utilize the validation set proposed in the Harmbench framework \cite{mazeika2024harmbench}, which contains 40 textual harmful behaviors across 7 semantic categories. For images, we use 5 adversarial images generated by an optimization-based jailbreak attack \citet{qi2024visual} and 5 clean images from ImageNet \cite{deng2009imagenet}. As a result, we have 200 adversarial text-image pairs and 200 clean pairs.

\vspace{-.5em}
\subsection{Findings}
According to the results displayed in Figure \ref{fig:findings}, the key findings can be summarized as follows:
\begin{figure*}[ht]
    \centering
    \captionsetup[subfigure]{justification=centering, margin=0pt, skip=0pt} 
    \begin{subfigure}{.45\textwidth}
        \centering
        \includegraphics[width=\textwidth]{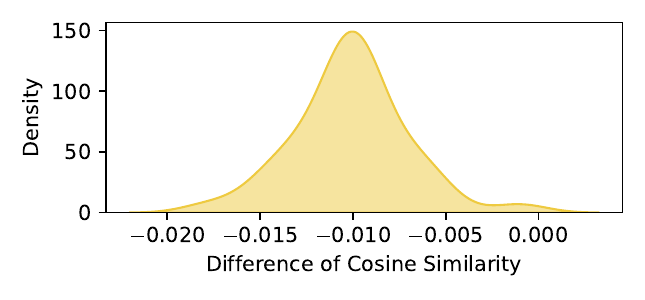}
        \vspace{-15pt}
        \caption{$\langle \bm{E^\textit{M}_{\textit{text}}}, \bm{E^\textit{C}_{\textit{img(o)}}} \rangle-\langle \bm{E^\textit{M}_{\textit{text}}}, \bm{E^\textit{A}_{\textit{img(o)}}} \rangle$}
        \label{fig:clean_adv_difference}
    \end{subfigure}
    \hspace{10pt}
    \begin{subfigure}{.45\textwidth}
        \centering
        \includegraphics[width=\textwidth]{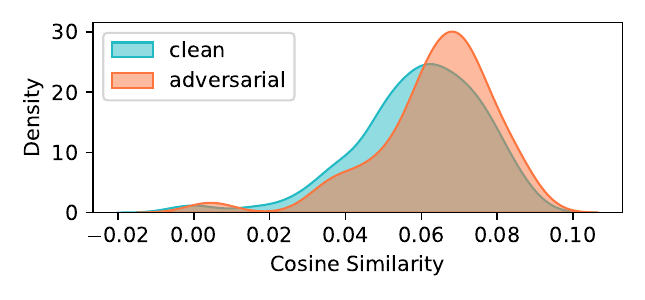}
        \vspace{-15pt} 
        \caption{$\langle \bm{E^\textit{M}_{\textit{text}}}, \bm{E_{\textit{img(o)}}} \rangle$}
        \label{fig:cos-sim}
    \end{subfigure}
    \begin{subfigure}{.45\textwidth}
        \centering
        \includegraphics[width=\textwidth]{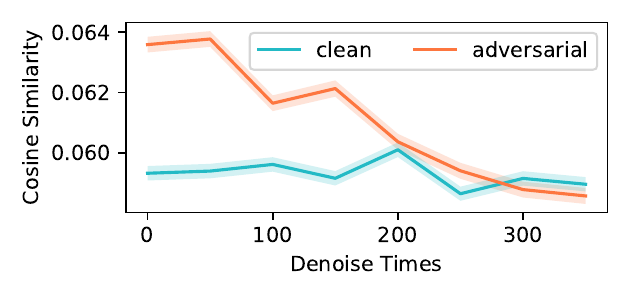}
        \vspace{-15pt}
        \caption{$\langle \bm{E^\textit{M}_{\textit{text}}}, \bm{E_{\textit{img(d)}}} \rangle$}
        \label{fig:cos-sim-trends}
    \end{subfigure}
    \hspace{10pt}
    \begin{subfigure}{.45\textwidth}
        \centering
        \includegraphics[width=\textwidth]{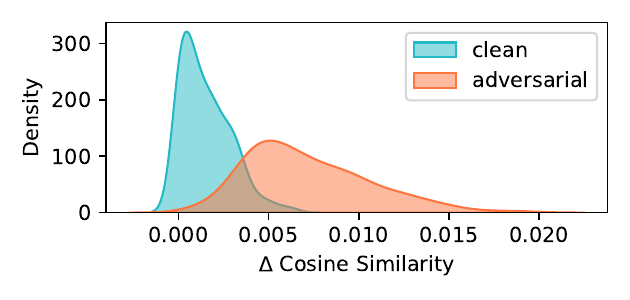}
        \vspace{-15pt}
        \caption{$\langle \bm{E^\textit{M}_{\textit{text}}}, \bm{E_{\textit{img(o)}}} \rangle - \langle \bm{E^\textit{M}_{\textit{text}}}, \bm{E_{\textit{img(d)}}} \rangle$}
        \label{fig:delta-cos-sim}
    \end{subfigure}
    \vspace{-5pt}
    \caption{Experimental result. (a) The distribution of the difference between clean and adversarial images regarding their cos-sim with harmful queries. (b) The distribution of cos-sim between harmful queries and clean/adversarial images. (c) The change of the cos-sim during denoising. (d) The distribution of $\Delta$cos-sim before and after denoising of clean/adversarial images.}
    \label{fig:findings}
\vspace{-1em}
\end{figure*}
\vspace{-.2em}
\subsubsection*{Finding 1: Adversarial images indeed contain harmful information.}
\vspace{-.2em}
For each harmful query, we calculate the cosine similarity between the queries and both clean and adversarial images,  denoted as $\langle \bm{E^M_{\textit{text}}}, \bm{E^C_{\textit{img(o)}}} \rangle$ and  $ \langle \bm{E^M_{\textit{text}}}, \bm{E^A_{\textit{img(o)}}} \rangle $ respectively. Figure \ref{fig:clean_adv_difference} shows the distribution of $\langle \bm{E^M_{\textit{text}}}, \bm{E^C_{\textit{img(o)}}} \rangle -\langle \bm{E^M_{\textit{text}}}, \bm{E^A_{\textit{img(o)}}} \rangle$. It can be observed that the distribution is almost entirely concentrated in the negative region, indicating that, for a specific harmful query, the semantic distance between it and an adversarial image is smaller than that between it and a clean image. Therefore, we can conclude that adversarial images indeed carry harmful information from queries.

\vspace{-.2em}
\subsubsection*{Finding 2: Directly utilizing the semantic difference between clean and adversarial images to harmful query is challenging}
\vspace{-.2em}
Figure \ref{fig:cos-sim} shows the distribution of the absolute value of $\langle \bm{E^M_{\textit{text}}}, \bm{E^C_{\textit{img(o)}}} \rangle$ and  $ \langle \bm{E^M_{\textit{text}}}, \bm{E^A_{\textit{img(o)}}} \rangle $. Although the distribution differs in the peak and concentration, distinguishing between adversarial and clean images based solely on the absolute value of the difference is challenging. This difficulty arises because the cosine similarity between different queries and adversarial images varies significantly, and the absolute value of the difference does not fully capture the characteristics of the images.

\vspace{-.2em}
\subsubsection*{Finding 3: Denoising can reduce harmful information but cannot eliminate}
\vspace{-.2em}
Subsequently, we applied denoising to each image 350 times, assessing cosine similarity with harmful queries every 50 iterations (visualization of the denoising is relegated to Appendix \ref{app:denoising}). Figure \ref{fig:cos-sim-trends} illustrates how cosine similarity between harmful query and adversarial images decreases as denoising progresses, indicating a reduction in harmful information. Despite this reduction, denoised adversarial images, when tested with harmful text inputs in the MLLM, still enabled a significant number of queries to jailbreak. Thus, while denoising mitigates harmful information in images, it does not eliminate their adversarial properties. 
\vspace{-.2em}
\subsubsection*{Finding 4: Relative shift in the semantic distance before and after denoising can help detect adversarial images.}
\vspace{-.2em}
In Figure \ref{fig:delta-cos-sim}, we present the change of cosine similarity before and after denoising ($\langle \bm{E^\textit{M}_{\textit{text}}}, \bm{E_{\textit{img(o)}}} \rangle - \langle \bm{E^\textit{M}_{\textit{text}}}, \bm{E_{\textit{img(d)}}} \rangle$). The distribution of cosine similarity between harmful queries and images shifts notably after denoising, contrasting with the distribution in Figure \ref{fig:cos-sim}. This observed shift supports our intuition that relative changes before and after denoising, rather than absolute differences, provide a method to distinguish adversarial images.  Figure \ref{fig:cos-sim-trends} further demonstrates this trend, showing a consistent decrease in cosine similarity between adversarial images and harmful queries, while the cosine similarity with clean images varies less.
\vspace{-.2em}
\section{Method}
\vspace{-.2em}
In this section, we first overview the defense pipeline and the components of \textit{CIDER}, followed by introducing the threshold selection strategy.
\subsection{Overview}
\textit{CIDER} is implemented on top of the MLLMs to defense optimization-based adversarial jailbreak attacks. Figure \ref{fig:workflow} presents the overview of the \textit{CIDER} pipeline. 
Specifically, given a text-image input pair, denoted as <$\textit{text}$, $\textit{img(o)}$>, \textit{CIDER} calculates the embeddings of text and image modalities, denoted as $\bm{E_{\textit{text}}}$ and $\bm{E_{\textit{img(o)}}}$. Then, the built-in denoiser in \textit{CIDER} will perform 350 denoising iterations on the image(o), calculating the denoised embeddings every 50 iterations, denoted as $\mathcal{E}=\bm{E_{\textit{img(d)}}}$. 
The $\textit{img(o)}$ will be identified as an adversarial example if any $\bm{E_{\textit{img(d)}}} \in \mathcal{E}$ satisfy the following condition: 
\vspace{-.5em}
\begin{align}
\langle \bm{E_{\textit{text}}}, \bm{E_{\textit{img(o)}}} \rangle - \langle \bm{E_{\textit{text}}}, \bm{E_{\textit{img(d)}}} \rangle >\tau 
\end{align}
where $\langle \cdot \rangle$ represents the cosine similarity and $\tau$ is the predefined threshold. Consequently, \textit{CIDER} will directly refuse to follow the user's request by responding ``I'm sorry, but I can not assist.'' if the image modality is detected as adversarial. Otherwise, the original image and query will be fed into the MLLM. The pseudo-code of \textit{CIDER} is illustrated in Algorithm \ref{code:detect}.

\vspace{-1em}
\begin{algorithm}\small
\caption{\textit{CIDER} defense pipeline}\label{code:detect}
\KwIn{
    $img(o)$: input image, $text$: input query, $\mathcal{F}$: target MLLM, $\tau$: predefined threshold.
}
$flag\gets true$\;
\For{$i \gets 0$ \KwTo $350$ \rm{\textbf{Step}} $50$}{
    $img(d)\gets denoiser(img(o),i)$\;
    $\bm{E_{\textit{text}}}\gets TextEncoder(text)$\;
    $\bm{E_{\textit{img(o)}}}\gets ImgEncoder(img(o))$\;
    $\bm{E_{\textit{img(d)}}}\gets ImgEncoder(img(d))$\;
    $d\gets\langle \bm{E_{\textit{text}}}, \bm{E_{\textit{img(o)}}} \rangle - \langle \bm{E_{\textit{text}}}, \bm{E_{\textit{img(d)}}} \rangle$\;
    \If{$d>\tau$}{
        $flag\gets false$\;
    }
}
\eIf{$flag=true$}{
    Return $\mathcal{F}(img(o),text)$\;
}{
    Return "I’m sorry, but I can not assist."
}
\vspace{-.5em}
\end{algorithm}
\vspace{-1.5em}

\vspace{-.2em}
\subsection{Threshold selection}

The threshold is selected based on the harmful queries and clean images ensuring that the vast majority of clean images pass the detection. The selection of threshold $\tau$ can be formulated as: 
\begin{small}
\begin{equation}
r=\frac{\sum\mathbb{I}(\langle \bm{E^\textit{M}_{\textit{text}}}, \bm{E^\textit{C}_{\textit{img(o)}}} \rangle - \langle \bm{E^\textit{M}_{\textit{text}}}, \bm{E^\textit{C}_{\textit{img(d)}}} \rangle <\tau) }{\# \textit{samples}} 
\end{equation}
\end{small}
where $r$ represents the passing rate and $\bm{E^\textit{M}_{\textit{text}}}$, $\bm{E^\textit{C}_{\textit{img(o)}}}$, $\bm{E^\textit{C}_{\textit{img(d)}}}$ stand for the embeddings of input query, the input image and denoised image respectively. The threshold $\tau$ is determined by controlling the passing rate $r$. For example, using the $\tau$ when setting $r$ to 95\% as the threshold indicates allowing 95\% percent of clean images to pass the detection.

The selection of the threshold significantly impacts the effectiveness of \textit{CIDER}: a threshold that is too high will cause many adversarial examples to be classified as clean samples, resulting in a low true positive rate (TPR); conversely, a threshold that is too low will lead to a high false positive rate (FPR), affecting the model's normal performance. 

\vspace{-1em}
\begin{figure}[ht]
\setlength{\abovecaptionskip}{0.2cm}
  \centering
  \includegraphics[width=.8\linewidth]{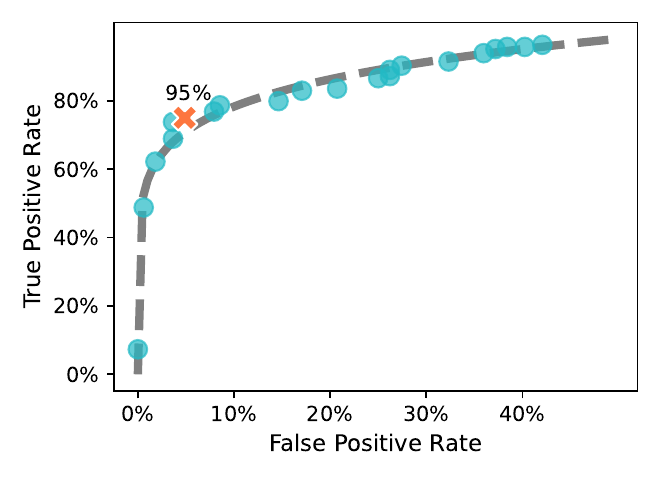}
  \vspace{-.5em}
  \caption{TPR-FPR trade-off on validation set.}
  \label{fig:tpr-fpr}
\vspace{-.5em}
\end{figure}

The ablation study is conducted to determine the optimal threshold. By treating adversarial pairs as positive samples and clean pairs as negative samples, we plot the TPR-FPR curve with thresholds ranging from 80\% to 100\% in 1\% increments, as shown in Figure \ref{fig:tpr-fpr}. Ideally, we expect high TPR and low FPR (the upper left corner of the plot). Therefore, we selected $\tau$ when $r$ equals 95\% as the detection threshold of \textit{CIDER}.
\vspace{-.2em}
\section{Experiment}
\vspace{-.5em}
In this section, we begin by outlining the configurations of our experiments, including the models, datasets, baselines, and evaluation metrics. We then evaluate the effectiveness and efficiency of \textit{CIDER}, comparing with the baseline methods. Next, we discuss the trade-off between robustness and utility, and the choice of denoising method. Finally, we demonstrate the generalization of our method.
\vspace{-.2em}
\subsection{Configurations}

\partitle{Models}
Note that \textit{CIDER} is an auxiliary model that is independent of the MLLMs. We use LLaVA to capture the semantic meaning of each modality, but \textit{CIDER} can be plugged into any other MLLMs. To demonstrate the effectiveness of \textit{CIDER}, we test the detection and defense performance on four open-source MLLMs, including LLaVA-v1.5-7B \cite{llava}, MiniGPT4 \cite{zhu2023minigpt}, InstructBLIP \cite{dai2024instructblip}, and Qwen-VL \cite{bai2023qwen}, as well as the API-access MLLM, GPT4V \cite{achiam2023gpt}. 

\begin{figure*}[]
\setlength{\abovecaptionskip}{0.2cm}
  \centering
  \includegraphics[width=.8\linewidth]{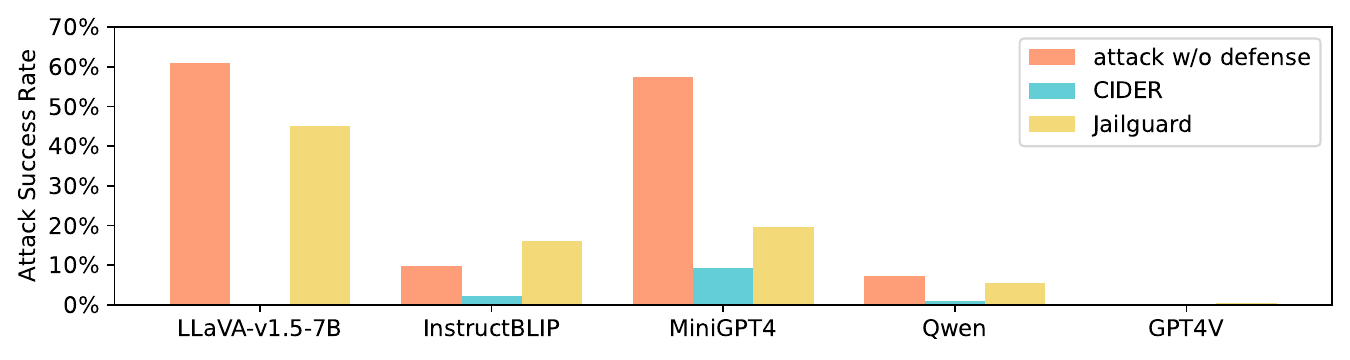}
  \vspace{-.5em}
  \caption{ASR of base MLLM, defending with \textit{CIDER} and defending with \textit{Jailguard}}
  \label{fig:asr}
   \vspace{-1em} 
\end{figure*}

\partitle{Datasets} Similar to the dataset used in Section \ref{sec:pre-experiment}, we generate 800 adversarial text-image pairs utilizing the 160 harmful queries in Harmbench \cite{mazeika2024harmbench} and adversarial images provided by \citet{qi2024visual}. To further demonstrate \textit{CIDER} will not destroy the original utilities on the normal queries, we also evaluate the utility of \textit{CIDER} protected MLLMs on MM-Vet benchmark \cite{yu2023mm}, which examines 6 core vision language capabilities, including recognition, optical character recognition (OCR), knowledge, language generation, spatial awareness, and math. 

\partitle{Baseline and evaluation metrics} We use \textit{Jailguard} \cite{zhang2023mutation} as a baseline, which is a SoTA mutation-based jailbreak detection strategy that protects the MLLMs at the inference stage. We involve four evaluation metrics to demonstrate the performance of defending methods from different aspects. From the perspective of the effectiveness of \textit{CIDER}, we incorporate detection success rate (\textbf{DSR}) and Attack success rate (\textbf{ASR}). \textbf{DSR} represents the proportion of adversarial examples $\mathcal{D}$ that can be successfully detected:

\begin{small}
\begin{equation}
    \text{DSR} \overset{\text{def}}{=} \frac{1}{|\mathcal{D}|} \sum_{(q,x_{adv})\in \mathcal{D}} \mathbb{I}_{adv}((q,x_{adv}))
\end{equation}
\end{small}

\textbf{ASR} is a standard evaluation metric indicating the proportion of samples that can successfully jailbreak MLLM $\mathcal{F}$ and generate harmful contents, which can be stated as: 

\begin{small}
\begin{equation}
\text{ASR} \overset{\text{def}}{=} \frac{1}{|\mathcal{D}|} \sum_{(q,x_{adv})\in \mathcal{D}} \mathbb{I}_{harm}(\mathcal{G}(\mathcal{F}(q,x_{adv})))
\end{equation}
\end{small}
$\mathcal{G}$ refers to an LLM classifier \cite{mazeika2024harmbench} that determines the harmfulness of a response. $\mathbb{I}_{adv}$ and $\mathbb{I}_{harm}$ represent the adversarial and harmful indicator. In terms of efficiency, we measure the time cost of processing a single query. In addition, to evaluate the model utility on regular tasks, and responses, we incorporate an online evaluator \cite{huggingfaceMMVetEvaluator} provided along with MM-Vet benchmark, which utilizes GPT-4 to generate a soft grading score from 0 to 1 for each answer.

\vspace{-.2em}
\subsection{Effectiveness}
\partitle{DSR}
We first demonstrate the overall DSR that \textit{CIDER} can achieve and compare it with the baseline method, \textit{Jailguard}. Table \ref{tab:dsr} shows that \textit{CIDER} achieves a DSR of approximately 80\%, while the DSR of \textit{Jailguard} varies, depending on the target MLLMs. Note that \textit{CIDER} is independent of the MLLMs, thus the DSR does not vary with the choice of MLLMs. However, \textit{Jailguard}'s detection capability relies heavily on the model's safety alignment, so the DSR also varies. MLLMs with good alignment achieve high DSR (e.g., GPT4V), while poorly aligned MLLMs have relatively low DSR (e.g., InstructBLIP). In other words, \textit{Jailguard} does not significantly enhance MLLM robustness against adversarial jailbreak attacks, whereas \textit{CIDER} does. Nonetheless, \textit{CIDER} achieves a higher DSR than most of the \textit{Jailguard} results, except \textit{Jailguard} on GPT4V.

\begin{table}[ht]
\begin{center}
\resizebox{\columnwidth}{!}{
\begin{tabular}{l|c}
\toprule[1pt]
Method & detection success rate ($\uparrow$)\\
\midrule
\text{\textit{Jailguard} with LLaVA-v1.5-7B} &$39.50\%$ \\
\text{\textit{Jailguard} with InstructBLIP} & $32.25\%$ \\
\text{\textit{Jailguard} with MiniGPT4} & $69.50\%$ \\
\text{\textit{Jailguard} with Qwen-VL} & $77.50\%$ \\
\text{\textit{Jailguard} with GPT4V}& $94.00\%$ \\
\midrule
\textbf{\textit{CIDER}} & $79.69\%$\\
\bottomrule[1pt]
\end{tabular}
}
\vspace{-0.5em}
\caption{DSR of \textit{CIDER} and \textit{Jailguard}}
\label{tab:dsr}
\end{center}
\vspace{-1.5em}
\end{table}

\partitle{ASR}
To evaluate the effectiveness of \textit{CIDER}, we measure the decline in ASR after applying \textit{CIDER}. Figure \ref{fig:asr} compares the original ASR without defense (red bar), ASR after \textit{CIDER} (blue bar), and ASR after \textit{Jailguard} (yellow bar). Note that, \textit{Jailguard} is solely designed to detect jailbreak input. To ensure a fair comparison, we add an output module following \textit{Jailguard}'s detection. Specifically, if \textit{Jailguard} detects a jailbreak, it will refuse to respond, similar to \textit{CIDER}. Otherwise, the MLLM will process the original input.

Across all models, defending with \textit{CIDER} significantly reduces the ASR, yielding better results than the baseline. This indicates that \textit{CIDER} effectively enhances the robustness of MLLMs against optimization-based jailbreak attacks. The most notable improvements are seen in LLaVA-v1.5-7B, where ASR drops from 60\% to 0\%, and in MiniGPT4, from 57\% to 9\%. For MLLMs with initially low ASRs, such as InstructBLIP and Qwen-VL, ASR is reduced to approximately 2\% and 1\% respectively. Another notable disadvantage of \textit{Jailguard} is observed in models like GPT4V, InstructBLIP, and Qwen-VL, which already had strong safety alignment and resistance to adversarial attacks. In these cases, the use of \textit{Jailguard} resulted in a slight increase in ASR.

We conclude that the threshold determined by \textit{CIDER} can be effectively applied to different MLLMs due to their shared transformer-based LLM backbones, which generate comparable representations of harmful information. This harmful information, distilled from malicious queries, is embedded into adversarial images using similar optimization-based attacks. As a result, the consistent noise patterns produced by these attacks across different MLLMs can be detected using the same threshold, highlighting the robustness and transferability of \textit{CIDER}.
\vspace{-.2em}
\subsection{Efficiency}

Timely inference is crucial for safeguarding MLLMs in real-world applications. Table \ref{tab:time} shows the time required to process a single input pair and generate up to 300 tokens with different MLLMs, comparing no defense, \textit{CIDER}, and \textit{Jailguard}.

\begin{table}[ht]
\vspace{-.2em}
\begin{center}
\resizebox{1\columnwidth}{!}{
\begin{tabular}{l|ccc}
\toprule[1pt]
Model & Original &  \textit{CIDER} & \textit{Jailguard}\\
\midrule
\text{LLaVA-v1.5-7B} & $6.39s$ &  $7.41s$ ($1.13\times$) & $53.21s$ ($8.32\times$)\\
\text{InstructBLIP} & $5.46s$ &  $6.48s$ ($1.22\times$) & $47.83s$ ($8.76\times$)\\
\text{MiniGPT4} & $37.00s$ &  $38.02s$ ($1.03\times$) & $313.78s$ ($8.48\times$)\\
\text{Qwen-VL} & $6.02s$ &  $7.04s$ ($1.19\times$) & $48.48s$ ($8.05\times$)\\
\text{GPT4V}& $7.55s$ &  $8.57s$ ($1.16\times$) & $61.04s$ ($8.08\times$)\\
\bottomrule[1pt]
\end{tabular}
}
\vspace{-0.5em}
\caption{Time cost to process a single pair of inputs.}
\label{tab:time}
\vspace{-1.5em}
\end{center}
\end{table}

\textit{CIDER} surpasses \textit{Jailguard} in efficiency, adding only 1.02 seconds per input pair on average, which is relatively acceptable compared to the original inference time. In contrast, \textit{Jailguard} requires 8-9 times the original processing time. Additionally, \textit{CIDER} detection is irrelevant to the number of generated tokens in the query answers. Therefore, \textit{CIDER} does not cause additional overhead when increasing the number of generated tokens, ensuring the stability of \textit{CIDER}'s efficiency.
\vspace{-.2em}
\subsection{Robustness-utility trade-off}
\begin{figure*}[ht]
\setlength{\abovecaptionskip}{0.2cm}
  \centering
  \includegraphics[width=.9\linewidth]{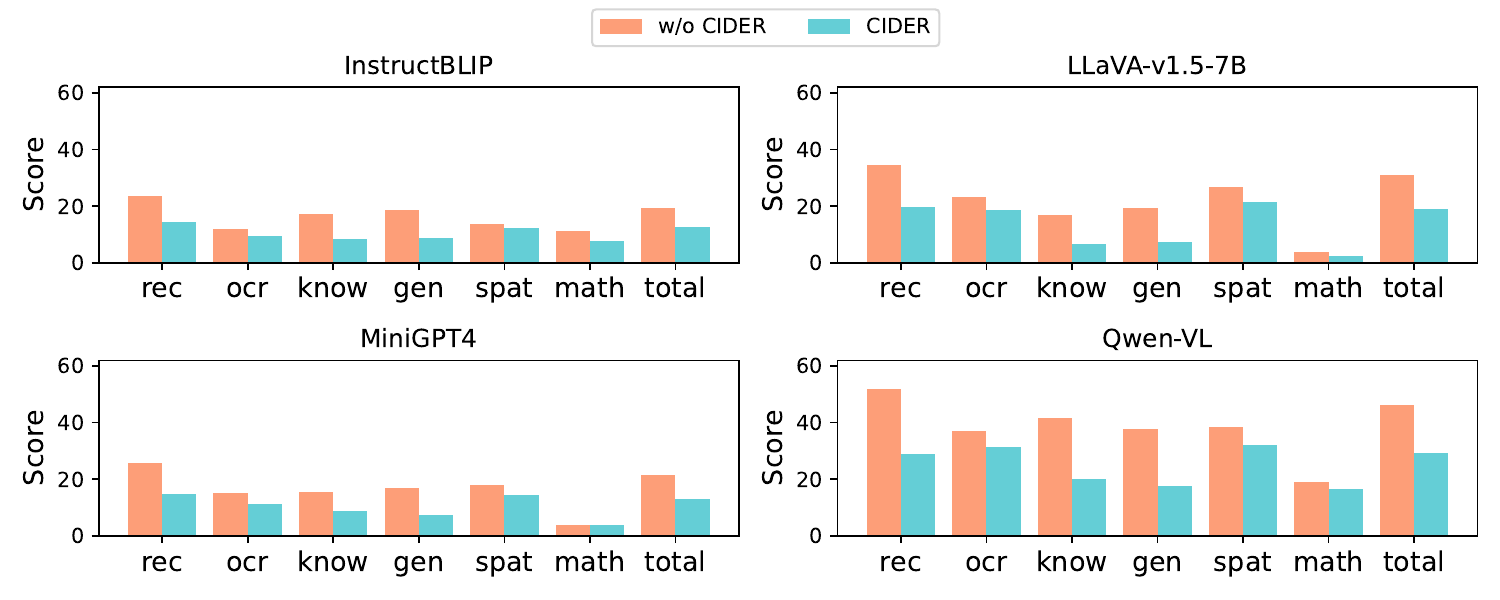}
  \vspace{-.5em}
  \caption{MLLM performance with and without \textit{CIDER} on MM-Vet.}
  \label{fig:mmvet}
  \vspace{-1em}
\end{figure*}

To further demonstrate \textit{CIDER}'s influence on the original utilities on normal queries, we also evaluate the utility of \textit{CIDER} protected MLLMs on MM-Vet benchmark, including recognition, OCR, knowledge, language generation, spatial awareness, and math. As shown in Figure \ref{fig:mmvet}, employing \textit{CIDER} leads to an approximate 30\% overall performance decline on normal tasks. Specifically, \textit{CIDER} mostly affects the MLLM's recognition, knowledge, and language generation capabilities, while it has minimal impact on OCR, spatial awareness, and math skills. We hypothesize that \textit{CIDER}'s stringent decision-making process, which outright rejects tasks once an image is identified as adversarial, hampers the model's overall performance. To further illustrate the robustness-utility trade-off, we conducted an ablation study using denoised images as inputs for the adversarial images, termed \textit{CIDER-de}. The result is relegated to Appendix \ref{app:cider-de}.

To find the optimal balance between safety and utility, we could design a more flexible rejection strategy, such as implementing multi-level thresholds for different types of content. This approach could reduce the negative impact on the model's functionality and we leave it to our future work.

\vspace{-.2em}
\subsection{Ablation study on denoising method}
We perform an ablation study on the choice of denoising method in the \textit{CIDER} architecture, as it significantly impacts both defense effectiveness and general task performance. Image smoothing methods commonly fall into two categories: DNN-based approaches, such as diffusion-based denoisers \cite{nichol2021improved}, and traditional filtering methods, like the Non-local Means filter (NLM; \citealp{buades2011non}).
We compare the ASR and MM-Vet scores of these two denoising methods, with the results presented in Table \ref{tab:denoiser}. The NLM filter performs similarly to the diffusion-based denoiser in terms of ASR, except on the LLaVA model, where it falls significantly behind. However, for general tasks, the NLM filter underperforms the diffusion-based denoiser across all models. This suggests that while the diffusion-based denoiser effectively reduces noise, it also preserves essential image details, making it a superior choice overall.

\begin{table}[htbp]

\centering
\resizebox{1\columnwidth}{!}{
\begin{tabular}{lcccc}
\toprule[1pt]
\multirow{2}{*}{Model}&\multicolumn{2}{c}{ASR($\%; \downarrow$)}& \multicolumn{2}{c}{Score($\uparrow$)}\\
\cline{2-5}
& Denoiser & NLM & Denoiser & NLM \\
\hline
\text{LLaVA-v1.5-7B} & $0.00$ & $8.12$ & $19.1$ & $18.9$\\
\text{InstructBLIP} & $2.34$ & $2.56$ & $12.7$ & $11.1$\\
\text{MiniGPT4} & $9.22$ & $9.06$ & $13.0$ & $11.2$\\
\text{Qwen-VL} & $1.09$ & $0.78$ & $29.2$ & $27.6$\\
\hline
\text{Average} & $\textbf{3.16}$ & $5.13$ & $\textbf{18.5}$ & $17.2$\\
\bottomrule[1pt]
\end{tabular}
}
\vspace{-0.5em}
\caption{The ASR and MM-Vet score of different denoising methods.}
\label{tab:denoiser}
\vspace{-1em}
\end{table}
\vspace{-.2em}
\subsection{Generalization}
\vspace{-0.2em}
In the previous sections, we evaluated the DSR and ASR against adversarial examples generated by \citet{qi2024visual}. To further assess the generalization of our defense method, which is critical for its applicability to other types of attacks, we test \textit{CIDER} under different attack settings.

\partitle{Dataset}
In addition to Harmbench, we employ RedTeam-2k \cite{luo2024jailbreakv} as a harmful query dataset, which compiles harmful queries from five different datasets and filters them according to the safety policies of OpenAI and Llama2, resulting in 2,000 red-teaming questions. We randomly select 200 queries and generate 800 text-image pairs using a similar processing method described in Section \ref{sec:pre-experiment}. On this dataset, \textit{CIDER} achieves a DSR of 81.37\%, with the reduction in ASRs presented in Table \ref{tab:transfer-dataset}. 
\begin{table}[htbp]
\vspace{-.3em}
\centering
\resizebox{.4\textwidth}{!}{
\begin{tabular}{lccc}
\toprule[1pt]
\multirow{2}{*}{Model}& Base& \multicolumn{2}{c}{\textit{CIDER}}\\
\cline{2-4}
& ASR(\%) & ASR(\%) & $\Delta$ (\%) \\
\hline
\text{LLaVA-v1.5-7B} & $29.87$ & $1.87$ & $28.00$\\
\text{InstructBLIP} & $24.13$ & $4.37$ & $19.76$\\
\text{MiniGPT4} & $43.75$ & $15.63$ & $28.12$\\
\text{Qwen-VL} & $10.62$ & $2.37$ & $8.25$\\
\bottomrule[1pt]
\end{tabular}
}
\vspace{-0.5em}
\caption{Generalization to RedTeam-2k dataset}
\label{tab:transfer-dataset}
\vspace{-1em}
\end{table}

\partitle{Attack Method}
To evaluate the generalization of our defense against different attack methods, we generated 800 adversarial pairs using ImgJP, an optimization-based approach proposed by \citet{niu2024jailbreaking}. Table \ref{tab:transfer-attack} shows the decrease in ASRs across four open-source MLLMs, with all ASRs falling below 4\%, and Qwen-VL achieving a 0\% ASR. Moreover, \textit{CIDER} attained a DSR of 93.87\% against ImgJP. These results demonstrate that \textit{CIDER} effectively generalizes to defend against optimization-based adversarial attacks, underscoring its practical value in real-world applications.
\begin{table}[htbp]
\vspace{-.3em}
\centering
\resizebox{.4\textwidth}{!}{
\begin{tabular}{lccc}
\toprule[1pt]
\multirow{2}{*}{Model}& Base& \multicolumn{2}{c}{\textit{CIDER}}\\
\cline{2-4}
& ASR(\%) & ASR(\%) & $\Delta$ (\%) \\
\hline
\text{LLaVA-v1.5-7B} & $61.0$ & $3.5$ & $57.5$\\
\text{InstructBLIP} & $4.0$ & $1.5$ & $2.5$\\
\text{MiniGPT4} & $52.5$ & $4.0$ & $48.5$\\
\text{Qwen-VL} & $6.5$ & $0.0$ & $6.5$\\
\bottomrule[1pt]
\end{tabular}
}
\vspace{-0.5em}
\caption{Generalization to ImgJP attack}
\label{tab:transfer-attack}
\vspace{-1em}
\end{table}

\vspace{-0.2em}
\section{Related Work}
\vspace{-0.5em}

\partitle{Multimodal Large Language Model}
A typical Multimodal Large Language Model (MLLM) consists of an image encoder \cite{dosovitskiy2020image} to extract feature maps, a projector to align image modality information with text modality, and a Large Language Model (LLM) to integrate textual and visual input for generating responses. The impressive multimodal capabilities of these models have spurred significant research interest, leading to contributions from both academia and industry \cite{achiam2023gpt, llava, zhu2023minigpt, dai2024instructblip, bai2023qwen}.

\partitle{Jailbreaking MLLMs}
Incorporating visual information into the LLM framework significantly broadens its range of applications but also introduces new security vulnerabilities, complicating the security issues of MLLMs. Besides transferring text jailbreak templates from LLMs to MLLMs \cite{luo2024jailbreakv}, effective strategies for jailbreaking MLLMs include using gradient-based methods to generate adversarial images \cite{carlini2024aligned, zhao2024evaluating, qi2024visual, niu2024jailbreaking}, and submitting screenshots containing harmful instructions \cite{gong2023figstep} or related images \cite{liu2023query, shayegani2023jailbreak}. This paper focuses on safeguarding MLLMs against gradient-based jailbreak attacks via adversarial images, aiming to fortify MLLMs against such sophisticated threats and ensure their robustness and reliability in practical applications.

\partitle{Safeguarding MLLMs}
Various defense mechanisms have been proposed to address vulnerabilities in MLLMs and enhance their security and robustness. These mechanisms can be categorized into proactive and reactive defenses based on their preventive and responsive nature. Proactive defenses aim to prevent attacks through techniques like adversarial training \cite{zong2024safety} and reinforcement learning \cite{chen2023dress} during the training phase. In contrast, reactive defenses focus on detecting attacks during the inference phase using methods such as \cite{wang2024inferaligner, pi2024mllm, wang2024adashield}. However, many of these methods require access to internal model parameters or rely on additional large models for implementation. Our approach prioritizes a reactive defense strategy for its practicality and ease of implementation. Notably, \textit{Jailguard} \cite{zhang2023mutation} is closely related to our work, as it detects jailbreak queries by analyzing variations in responses to perturbed inputs. However, \textit{Jailguard}'s detection success heavily depends on the safety of the underlying LLM and involves significant computational costs.

\vspace{-.5em}
\section{Conclusion}
\vspace{-.5em}
In this work, we propose a plug-and-play cross-modality information detector, \textit{CIDER}, which can effectively and efficiently defend against adversarial jailbreak attacks. Compared to previous methods, \textit{CIDER} achieves superior defense performance, as evidenced by higher DSR and a significant decline in ASR, while greatly reducing processing time. We also evaluate the generalization of \textit{CIDER} to other datasets and optimization-based adversarial attacks, and demonstrate the robustness-utility trade-off in MLLMs. In future research, we aim to improve \textit{CIDER} by reducing the negative impact on MLLM utilities to normal tasks. Additionally, it would be useful to develop defense mechanisms against non-optimization-based jailbreak attacks.

\section*{Acknowledgement}
\vspace{-.5em}
We thank all reviewers for their constructive comments. This work is supported by the Shanghai Engineering Research Center of Intelligent Vision and Imaging and the Open Research Fund of The State Key Laboratory of Blockchain and Data Security, Zhejiang University.

\section*{Limitations}
\vspace{-.2cm}
We outline the limitations of our study as follows:

1. While \textit{CIDER} is an effective, efficient, and user-friendly defense mechanism, it does impact MLLM performance to some extent. We believe this is due to \textit{CIDER}'s stringent handling of adversarial examples. In future work, we plan to implement multi-level thresholds to process adversarial examples with varying degrees of rigor, aiming to maintain robust defense capabilities without compromising MLLM performance.

2. \textit{CIDER} is specifically designed to defend against optimization-based adversarial jailbreak attacks, and its effectiveness against other types of jailbreak attacks is uncertain. Future research will explore \textit{CIDER}'s effectiveness against these alternative attacks and develop corresponding defense strategies, aiming to enhance the overall security and resilience of MLLMs against a wider array of adversarial threats.

\section*{Ethics Statement}
\vspace{-.2cm}
Ensuring the security of Vision Large Language Models (MLLMs) is crucial as they become more widely used in various applications. This paper introduces \textit{CIDER}, a simple yet effective cross-modality information detector designed to defend against adversarial jailbreak attacks in MLLMs. Our work significantly contributes to the field by providing a tool that mitigates known vulnerabilities and lays the groundwork for future improvements in safety measures.
While \textit{CIDER} marks significant progress, it doesn't make MLLMs immune to all threats. Continuous evaluation and updates are crucial as MLLMs evolve. By sharing \textit{CIDER} and our findings, we aim to encourage ongoing research and collaboration, promoting advanced and secure MLLMs.
We are committed to addressing the ethical implications of MLLM deployment, ensuring transparency, and prioritizing the responsible use of these technologies for societal benefit.

\vspace{-.5cm}

\bibliography{custom}

\appendix
\section{Visualization of denoising}\label{app:denoising}
\begin{figure}[ht]
\setlength{\abovecaptionskip}{0.2cm}
  \centering
  \includegraphics[width=\linewidth]{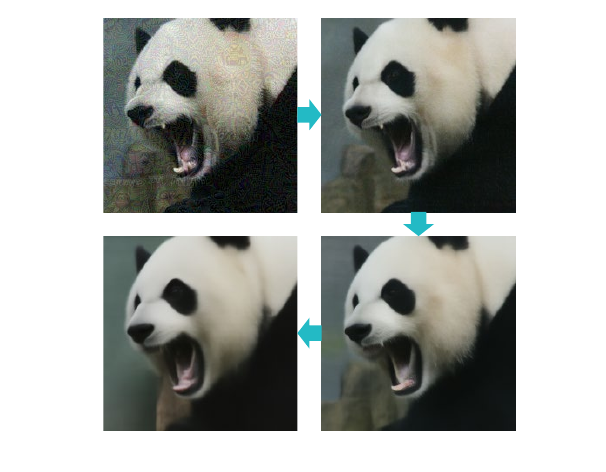}
  \caption{An example of the denoising procedure. }
  \label{fig:denoise-eg}
\end{figure}

Figure \ref{fig:denoise-eg} presents an example of an adversarially perturbed image, showing the effects of denoising it after 100, 200, and 300 iterations.

\section{Ablation study on robustness-utility trade-off}\label{app:cider-de}
\begin{figure*}[ht]
\setlength{\abovecaptionskip}{0.2cm}
  \centering
  \includegraphics[width=\linewidth]{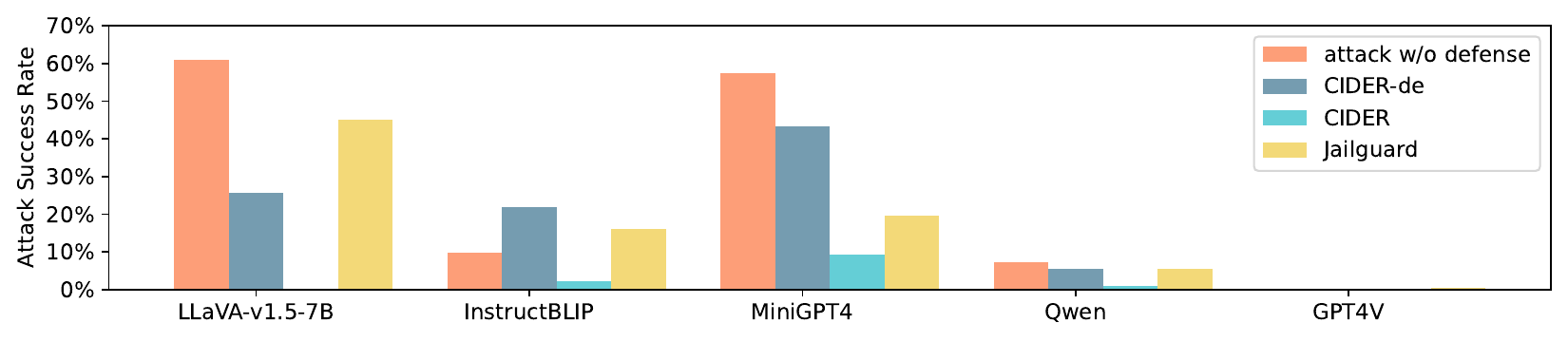}
  \caption{ASR of base MLLM, defending with \textit{CIDER-de}, \textit{CIDER} and \textit{Jailguard}}
  \label{fig:asr-full}
\end{figure*}

\begin{figure*}[ht]
\setlength{\abovecaptionskip}{0.2cm}
  \centering
  \includegraphics[width=.9\linewidth]{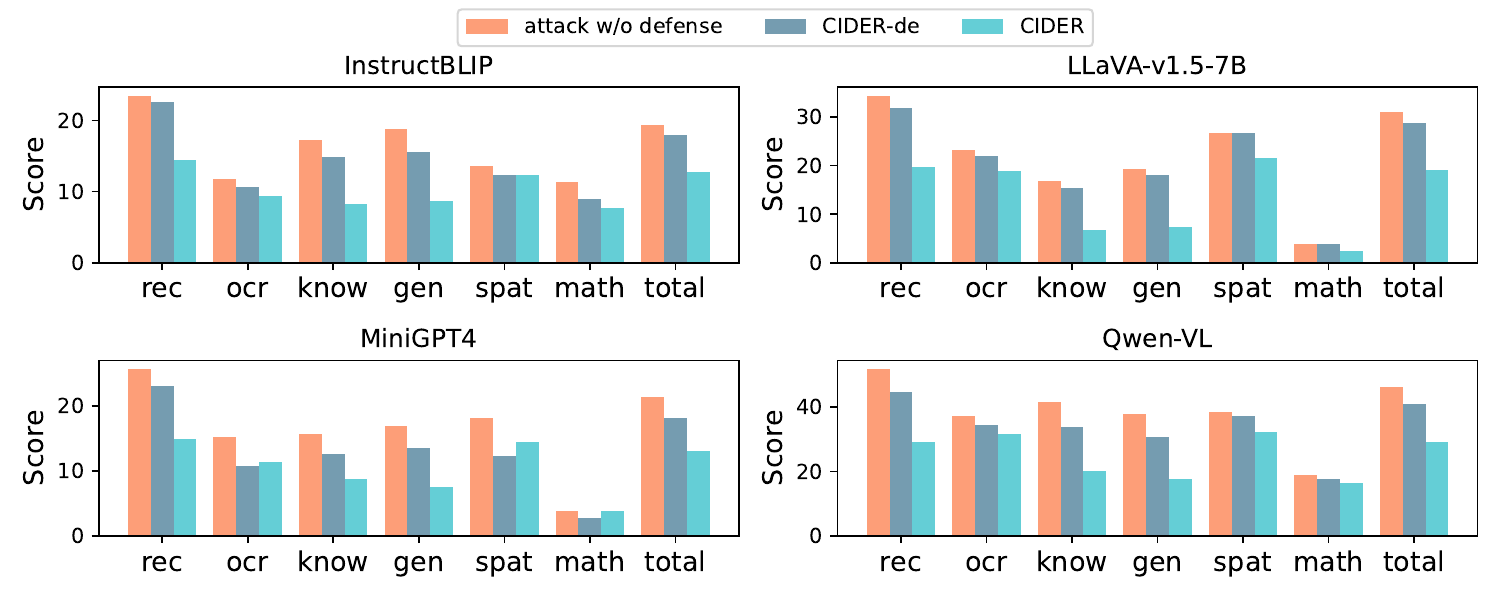}
  \caption{MM-Vet score of base MLLM, defending with \textit{CIDER-de} and \textit{CIDER}}
  \label{fig:mmvet-de}
\end{figure*}

To further illustrate the robustness-utility trade-off, we perform an ablation study using denoised images as inputs for adversarial images, referred to as \textit{CIDER-de}. Figure \ref{fig:asr-full} shows the ASR of \textit{CIDER-de} and Figure \ref{fig:mmvet-de} shows the MM-Vet score of it. It can be observed that using \textit{CIDER-de} hardly impacts the utility of the MLLM. However, this comes at the expense of greatly diminished defensive effectiveness.

\end{document}